# SENTIMENT ANALYSIS OF CYBER SECURITY CONTENT ON TWITTER AND REDDIT

Bipun Thapa

College of Business, Innovation,
Leadership, and Technology, Marymount University, USA

## ABSTRACT

*Sentiment Analysis provides an opportunity to understand the subject(s), especially in the digital age, due to an abundance of public data and effective algorithms. Cybersecurity is a subject where opinions are plentiful and differing in the public domain. This descriptive research analyzed cybersecurity content on Twitter and Reddit to measure its sentiment, positive or negative, or neutral. The data from Twitter and Reddit was amassed via technology-specific APIs during a selected timeframe to create datasets, which were then analyzed individually for their sentiment by VADER, an NLP (Natural Language Processing) algorithm. A random sample of cybersecurity content (ten tweets and posts) was also classified for sentiments by twenty human annotators to evaluate the performance of VADER. Cybersecurity content on Twitter was at least 48% positive, and Reddit was at least 26.5% positive. The positive or neutral content far outweighed negative sentiments across both platforms. When compared to human classification, which was considered the standard or source of truth, VADER produced 60% accuracy for Twitter and 70% for Reddit in assessing the sentiment; in other words, some agreement between algorithm and human classifiers. Overall, the goal was to explore an uninhibited research topic about cybersecurity sentiment.*

## 1. INTRODUCTION

Through rapid digitization across the globe that produces a voluminous amount of public data, mostly through social media [1], an area of research that is burgeoning is sentiment analysis or opinion mining [2]. The use of NLP [3] to understand the sentiment of public opinion often presents a prelude to a bigger picture, invaluable and prescriptive information in the digital age. Organizations, political parties, technology, and dependents to the public sentiment or opinion benefit from the foresight [4]. In this context, descriptive research on the perception of cybersecurity in the public domain would provide meaningful feedback to the industry and the entities involved [5]. Presumptively, cybersecurity is often viewed through cynical lenses, and with the frequent unfolding of negative events in the news media [6], it would be insightful to understand if a similar sentiment persists in the social media platforms. Leveraging Twitter data to analyze public sentiment in the modern research literature is common [7]; however, it is partial to another popular platform, Reddit, which has shown to be influential in its rights [8]. The research area of sentiment analysis is relatively young, less than 15 years, albeit experiencing a surging growth due to data availability, with most of the earlier studies focusing on the most optimum algorithms for classification [9]. The objectives of sentiment analysis could be for understanding customer feedback [10], perception of the healthcare system [11], or to improve education quality from an educator's point of view[12], among many other things. Similarly, this research is intended to focus on cybersecurity, which through initial analysis is an uninhibited domain in sentiment analysis and hence can potentially provide insights for actors involved. To conduct the research, social media content about cybersecurity, the topic, on Twitter and Reddit will be collected and analyzed through VADER (Valence Aware Dictionary for sEntiment Reasoning), a classification tool [13]. The documents from each source, Twitter and Reddit, will be collected to analyze the sentiment of the content. Concurrently, a small sample size of the content from Twitter and Reddit is human-classified for comparison. Human classification of the content could be utilitarian in understanding the correlation between machine-produced classification and human-produced classification [14], and to understand the efficacy of the algorithm. Sentiment, for this research, is labelled to be either 'positive', 'negative' or 'neutral', for both human and machine classification techniques. In essence, this research primarily aims to explore an inquisitive query, which is to know the opinion of cybersecurity through social media content for a specific time duration. In addition, the collected content will be preprocessed, which is to clean and normalize irregular content to yield better algorithm performance [15], and assessed for the sentiment classification. The classification evaluations metrics [16] will be used to measure the performance of the VADER against human labelling and frequenting entities in the content will be determined.

The three research objectives (RO) are,

RO1: To understand the sentiment of cybersecurity content posted on Twitter and Reddit in a given timeframe.
RO1.1: To understand the sentiment of cybersecurity content posted on Twitter and Reddit in a given time frame without text preprocessing.
RO2: To determine the most mentioned entities on Reddit and Twitter.
RO3: To evaluate VADER against human classification.

The methodology to fulfill the research objectives above are listed below.

## 2. SIGNIFICANCE, ASSUMPTIONS AND LIMITATIONS

This precedes explanatory research in sentiment analysis of Cybersecurity content on social media while assuming that human classification is accurately reflected for comparison. The limited sample size of human classifiers in the survey could have established biased opinions.

## 3. LITERATURE REVIEW

Often classifying a sentiment (positive, negative or neutral) of opinions is considered to be a difficult classification problem to solve even with the machine learning integration. The insistence on understanding, and modelling, to understand the sentiment of opinion, however, has not been abated; frequently new researches offering higher performance or unique approaches are presented. Digitization has yielded an enormous amount of data, and an abundance of publicly accessible communication mediums (mainly social media) allows for researchers to observe the pulse of the public sentiment. With this research area being fluid and ever-changing, for literature review, the focus was on peer-reviewed journal articles, conference papers, and API libraries, post-2010 with exception of an article from 2013. It was important to reference, learn and identify gaps from recent work due to the dynamic nature of the field. The literature review observed existing practice and their outcomes, positive and negative. Thereafter, novel methods of refining existing research ideas and methodologies were identified.

Regarding machine learning algorithms, two popular approaches are; supervised and unsupervised learning algorithms [17]. Supervised learning starts with labeled input data, which have defined features or attributes. With the features, algorithms are presented with the relationship of the data, then trained and asked to perform. Unsupervised learning is without the labeled input data; the algorithm will identify inherent structure or relationship that cannot be easily and manually replicated. Naïve Bayes, Decision Trees, and Support Vector Machine are examples of supervised learning which have been used to identify sentiments.

Whilst not a machine learning algorithm in the traditional sense, the Lexicon-Based Approach (LBA), VADER, provides an alternative to understanding sentiments; it has a compilation of previously classified sentiment terms that it will compare with and yield a polarity score to determine its sentiment. LBA is further divided into the Dictionary-Based Approach (DBA) and Corpus-Based Approach (CBA). DBA gathers a smaller set of words without context and classifies their polarity, making DBA somewhat flawed. CBA uses statistical or semantic tools to find context to the words, but it is not as nearly efficient in gathering the data and will take a much longer time to develop a large set [18].

Carlos Costa et al., to address their explanatory research objective of Portuguese parties (RO2: Identify the global sentiment per political party in Twitter communication), used rule-based VADER to classify the sentiment of the tweets. The tweets were first translated into English then assessed via VADER for polarity score, which then was aggregated by the party for comparative score [19].

U. Yaqub et al., conducted subjectivity and polarity analysis, sub-domain of sentiment analysis, on Twitter data of US and UK elections. As their exploration showed, the online sentiment in many ways reflected real public opinion. They postulate that being able to understand the online (Twitter) sentiment can greatly predict the public opinion or decision that is forthcoming regardless of differences in data collection [20]. A larger tweet dataset and inclusive study of all states (US), not just ten populous states, would be more beneficial to determine the outliers.
Another research tries to find a relationship, in the form of correlation, between the sentiment analysis of StockTwits, a microblogging site, and the stock price to accurately predict the direction of the price [21]. The researchers used supervised machine learning algorithms and featurization techniques with a positive correlation and accuracy of more than 61 percent in the five companies examined.

Due to sentiment analysis being a complex classifier problem, it has lagged behind in producing accuracy compared to categorization problems by almost 10 percent [22], and in order to yield higher results, preprocessing or cleaning of input data is important. Haddi et al. recommend data transformation/filtering, classifying, and evaluating. In the first stage, data is isolated from tags and removed stop words. For the second step, using the 4:1 ratio for training: testing with 10 folds cross-validation and lastly evaluating the performance of the model. Some or all combinations of these can greatly reduce the noise, which hinders producing strong sentiment analysis results.

Numerous researches have been conducted to find sentiment analysis of Twitter data using various algorithms and methodologies. Reddit, another popular social media outlet that is abreast with fresh news, is fairly uninhibited by the research community. The sentiment classification has been mostly done through automated computing without the correlation of results with human sentiment analysis to see if they differ significantly. The gap this research can attempt to fulfill, therefore, is an analysis of sentiments from the same data topic (cybersecurity) derived from Twitter and Reddit, which would then be classified by humans, labeling it with a sentiment. The latter will be inferential in nature, simply validating random classifications from the algorithm to find commonality or discord when compared against human classification

## 4. METHODS

A descriptive design is adopted for this research, which notes the observation of the phenomenon but will not be able to conclude why it is occurring [23]. In this research, social media contents can be classified but its causation cannot be confirmed by corroborating evidence. The content from social media, Twitter and Reddit, is collected, analyzed, and presented to provide insight into a research area currently uninhibited, potentially providing a precursor to deeper and narrower research in the future.

Figure 1 provides a general overview of the research methodology with data preprocessing. The research dataset is created by filtering Reddit (posts) and Twitter (tweets) for cybersecurity content that falls under specific criteria (Table 3). From those two platforms, two different comma-separated values (CSV) files, twitter.csv, and reddit.csv, are extracted using API libraries praw and tweepy. Those files are preprocessed to be legible for sentiment analysis and fed through the VADER (Valence Aware Dictionary for sEntiment Reasoning) algorithm, which classifies the content of the CSV file appropriately. A small sample of 10 items from each file, which have been analyzed by VADER is collected and presented to the human surveyors for annotation through the form of a survey. The VADER classification and human classification are then compared to find differences, if any.

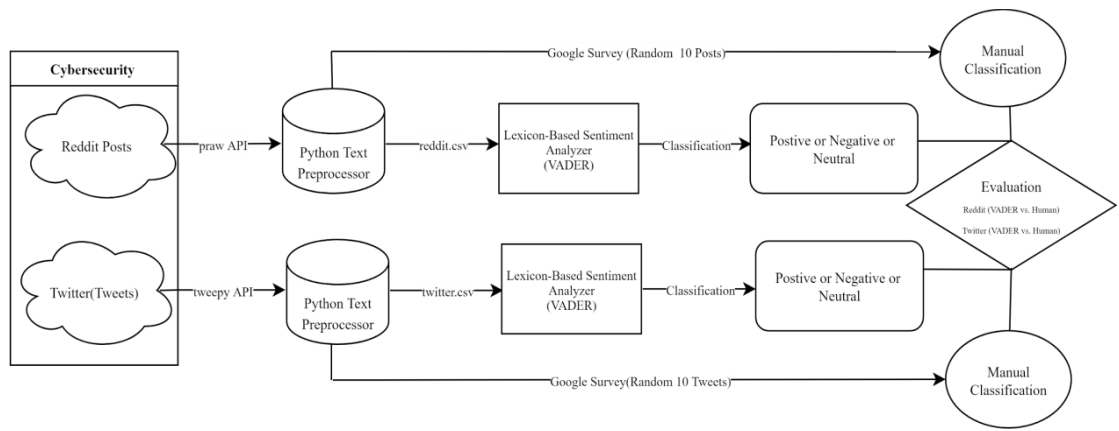

Figure 1. Proposed Method (With Python Text Preprocessing)

In Figure 2, text preprocessing is removed; as demonstrated by the creators of VADER [13], the extra step might not provide much value to the analysis of the sentiment.

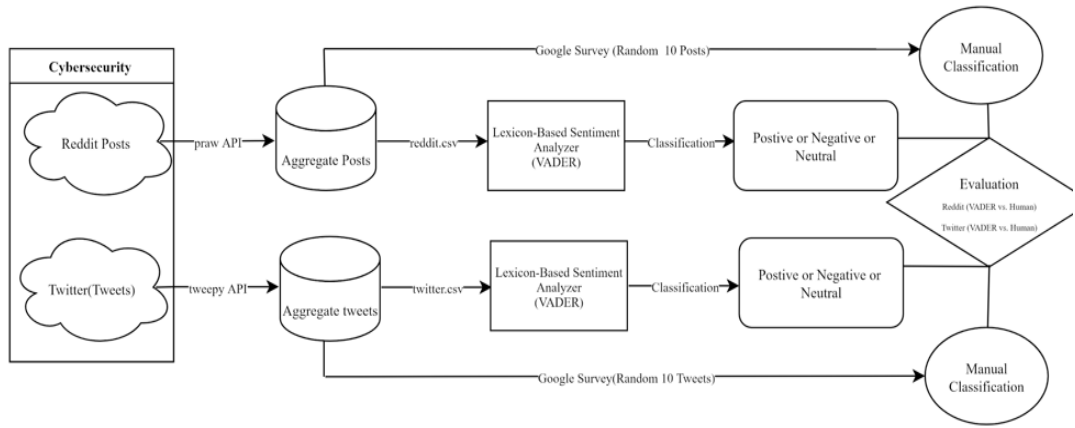

Figure 2. Proposed Method (Without Python Text Preprocessing)

## 4.1. PRAW (The Python Reddit API Wrapper)

The PRAW is a read-only API wrapper to extract Reddit posts through Python. It requires developer enrollment in Reddit, which provides credentials (Client ID, Client Secret, and User-Agent); this information authenticates the request against the Reddit server to gather posts from subreddits as desired. For this methodology, 'top' and 'hot' are gathered, which identify trending posts on Reddit[24]

## 4.2. Tweepy (Python Library for Accessing the Twitter API)

Similar to PRAW, Tweepy requires a developer application, when approved, will provide the users with credentials (consumer_key, consumer_secret,access_token, and access_token_secret) to authenticate and collect tweets depending on various parameters[25]. For this methodology, tweets with certain hashtags were collected.

## 4.3. Text Preprocessing

Text preprocessing is often recommended for data optimization when working with a large number of unstructured entries. The impetus is high on social media data that contains informal language, repetitions, URLs, and abbreviations [26], which creates noise, and as such clouds the 'real' sentiment behind the opinion. Figure 3 illustrates common steps for Text Preprocessing but can be expanded depending on the data type.

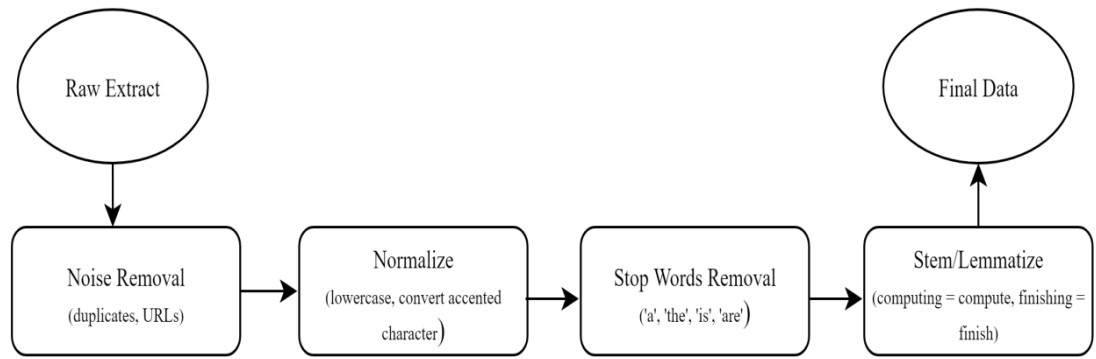

Figure 3. Standard Text Preprocessing for Sentiment Analysis

However, Hutto and Gilbert, the developers of VADER, state that sentiment heuristics play an important role in estimating the writer's mood. The punctuation, capitalization, tri-grams can amplify the mood [13]. Therefore, the dataset (reddit.csv and twitter.csv) for RO1.1 is directly fed into the VADER tool to understand the significance of

### 4.4. VADER (Valence Aware Dictionary for sEntiment Reasoning)

The foundation of this open-sourced, NTLK algorithm is a dictionary with corresponding sentiment features. The content, words, and phrases included are rated for polarity and intensity based on its built-in comprehension that recognizes more than 7500 features from -1 to +1, which the algorithm assesses for polarity, 'negative', 'positive, and 'neutral' scores, yielding a final 'compound' score. For a sentence, a final classification is based on 'compound' from the words. A positive compound score is 'positive', a negative 'compound' score is 'negative', and '0' is neutral. VADER's results in the past research have had outstanding accuracy at (F =0.96) compared to human classification, which was at (F1 = 0.84) [13], which makes this a popular tool of choice. Table 1 illustrates the VADER classification of individual words with an appropriate sentiment rating.

Table 1. VADER Sentiment Classification (Words)

| Word | Positive | Neutral | Negative | Compound | Final |
| --- | --- | --- | --- | --- | --- |
| 'sentiment' | 0 | 1 | 0 | 0 | Neutral |
| 'dangerous' | 0 | 0 | 1.0 | -0.47 | Negative |
| 'excellent' | 1 | 0 | 0 | 0.57 | Positive |

Table 2 classifies the sentiment of the entire sentence; similarly, a large data set can be processed in bulk to find the aggregate classification.

Table 2. VADER Sentiment Classification (Sentences)

| Word | Positive | Neutral | Negative | Compound | Final |
|---|---|---|---|---|---|
| 'data uses python.' | 0 | 1 | 0 | 0 | Neutral |
| 'sentiment is interesting' | 0.57 | 0.42 | 1.0 | 0.40 | Positive |
| 'security is difficult' | 0.40 | 0.16 | 0.42 | -0.02 | Negative |

### 4.5. Survey

There were thousands of tweets and posts collected via the framework to create the dataset, and it would not be possible to classify these manually. To find a comparative baseline, using Python's random.sample() function, ten tweets and ten Reddit posts were extracted. Using Google Form, the participants were asked to classify the tweets and posts into three sentiment classifications (positive, neutral, and negative). The survey was close-ended and anonymous, with participants required to classify the content with just one answer to each question. The survey adopted Snowball Sampling Method, where the form was posted on social media, and participants were requested to recruit other participants to increase the sampling size. The participants could originate from any sector, and upon completion of the survey, the human-classified sentiment could be used to be compared against VADER-generated analysis. The responses for each post and tweet would be assessed for polarity, with the majority score providing designation for final classification.

### 4.6. Data Extraction Criteria

The twitter.csv and reddit.csv datasets were created over a one-week window. Reddit has multiple subreddits (sections) dedicated to cybersecurity content; the three most popular ones were picked. Similarly, for tweets, the same hashtags bearing the name of the subreddits were picked for consistency. For Reddit, only posts that were 'hot' or 'top' were chosen, and on Twitter, tweets with at least one 'like' was chosen. This was to ensure that there was no favorability of the content by another user. Filtering using this method would substantially decrease the observations in the dataset.

Table 3. Dataset Creation Criteria

| | Window for Collection | Section from Social Media | Content Filtering |
|---|---|---|---|
| Reddit | 10/27/2021 - 03/11/2021 | Subreddits(Cybersecurity, computer security, privacy) | only 'top' or 'hot' posts |
| Twitter | 10/27/2021 - 03/11/2021 | Hashtags(#cybersecurity, #computersecurity, #privacy) | only tweets with likes >0 |
| Survey | 04/11-2021- 11/11/2021 | N/A | N/A |

## 5. RESULTS AND DISCUSSIONS

### 5.1. To Understand the Sentiment of Cybersecurity Content Posted on Twitter and Reddit in A Given Timeframe

The preprocessing of posts and tweets included removing stop words, stemming, normalizing, as shown in Figure 3. It is intended to clean up the noise, which potentially could alter the sentiment of the contents. There were 32481 and 1205 observations made from Twitter and Reddit, respectively within the aforementioned timeframe and criteria. From Table 4 below, the Base Polarity, which is the standard scale, indicates that there were more positive sentiments than negative, both in Twitter and Reddit Posts. The Moderate Polarity expands the classification criteria, where the content has to be over the .25 threshold (negative or positive); this is to identify firmer sentiments. Consequently, this increased the neutral distribution, but still, there were more positive sentiments on both platforms than negative. Lastly, Extreme Polarity measures strong sentiment toward the content where the polarity threshold was .75 (negative or positive). Again, positive sentiments supersede negative sentiments. Twitter was more conducive to positive sentiments than Reddit, albeit the latter has a significantly smaller sample size. Most Reddit posts were neutral, whereas Tweets were either positive or neutral. Cumulatively, the sentiment on cybersecurity content is mostly positive or neutral, while negative sentiments last in every assessment.

Table 4. Polarity Classification for Preprocessed Data

|  | Obs. | Base Polarity 0=neu,>0=pos,<0=neg | Moderate Polarity >0.25=pos,-.25 <0=neg | Extreme Polarity 0.75=pos,-.75 <0=neg |
|---|---|---|---|---|
| Twitter | 32481 | 49%= pos<br>28% = neu<br>22.5% = neg | 42%= pos<br>41% = neu<br>16.5% = neg | 8%= pos<br>90% = neu<br>1.7% = neg |
| Reddit | 1205 | 26.5%= pos<br>56% = neu<br>17% = neg | 22.5%= pos<br>64.5% = neu<br>13% = neg | 1%= pos<br>98.5% = neu<br>0.4% = neg |

*Note: Obs. = Observations, 'pos' = positive, 'neu' = neutral and 'neg' = negative*

### 5.2. To Understand the Sentiment of Cybersecurity Content Posted on Twitter and Reddit in a Given Time Frame without Text Preprocessing

Since the goal was to find the overall sentiment of the observations in percentage, duplicated observations would alter the representation; hence, one exception was made to raw data; only duplicates were removed. The standard preprocessing (Figure 3) was avoided and observations were directly fed in VADER to yield the polarity assessment. The unprocessed Twitter content produced similar sentiment scores compared to processed data; it had a similar narrative and the deviation between the comparable scores was less than 2%, with most of the sentiment being positive or neutral, indicating that preprocessing of Twitter content didn't shift the narrative. The Reddit posts were much more affected by the preprocessing; a significant percentage of neutral content became positive. A minor increase in negative scores also occurred but was not as significant. Overall, unprocessed content was in line with the narrative that most cybersecurity contents on these platforms were mostly positive or neutral, as listed in Table 5.

Table 5. Polarity Classification without Preprocessed Data

|  | Obs. | Base Polarity 0=neu,>0=pos,<0=neg | Moderate Polarity >0.25=pos,-.25 <0=neg | Extreme Polarity >0.75=pos,-.75 <0=neg |
|---|---|---|---|---|
| Twitter | 32481 | 48%= pos<br>29% = neu<br>22.5% = neg | 41.5%= pos<br>42% = neu<br>16.5% = neg | 8.5%= pos<br>89.5% = neu<br>1.6% = neg |
| Reddit | 1205 | 34%= pos<br>45% = neu<br>20% = neg | 30%= pos<br>55% = neu<br>15% = neg | 1.8%= pos<br>97% = neu<br>0.74% = neg |

Note: Obs. = Observations, 'pos' = positive, 'neu' = neutral and 'neg' = negative

## 5.3. To Determine the most Mentioned Entities on Reddit and Twitter

As per Merriam-Webster dictionary, an entity is defined as something with independent existence [27]. An entity for this objective could be a company, a specific product or technology, and a uniquely identifiable government. To address this, the two datasets were split into singular words, and the frequency was counted to identify the most discussed entities on Twitter and Reddit. Microsoft, Facebook, and Apple made it to the top ten list as the companies; GitHub, Node.js, and Chrome appeared as technologies, with the United States being the lone government. Figure 4 provides a graphical illustration by ranking.

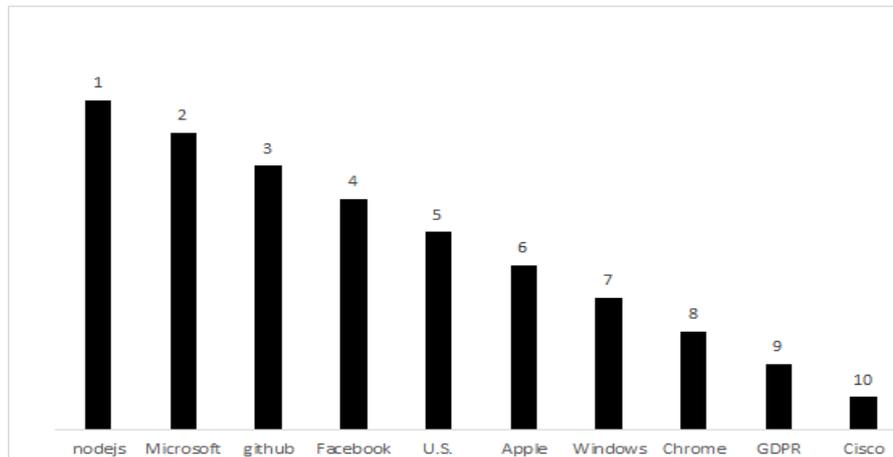

Figure 4. Most Discussed Entities on Twitter

Figure 5 lists the top ten entities from the Reddit cybersecurity posts by ranking. Reddit is most discussed, which could potentially be noise and not relevant to cybersecurity content. The big techs like Facebook, Microsoft, and Google are mentioned, along with the United Kingdom and China. Yubikey stands out as it is not a household name, but this could be due to its popularity during the data collection time.

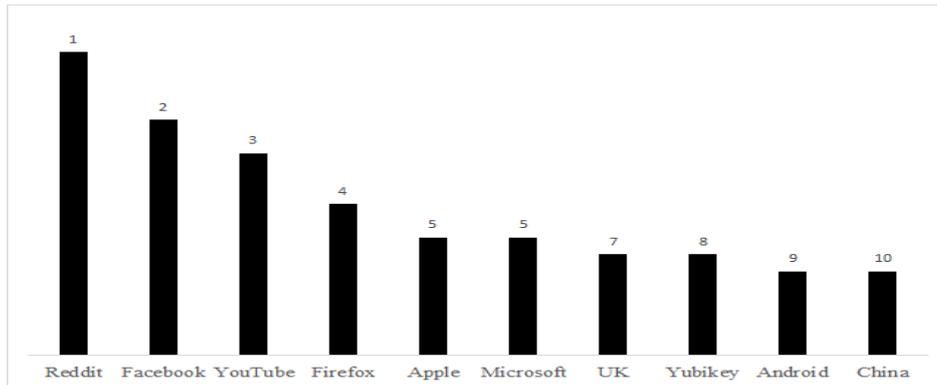

Figure 5. Most Discussed Entities on Reddit

The entities that appeared on Reddit and Twitter were, as expected, a combination of big technology firms, popular products, and governments that are at the forefront of cybersecurity development.

### 5.4. To Evaluate the VADER Algorithm against Human Classification

The survey conducted requested human participants to classify Twitter and Reddit contents. The majority classification ('neu', 'pos', or 'neg') of a post or a tweet to an appropriate polarity would be the baseline classification for VADER to be compared against. If the VADER assigned the same classification as human participants, the accuracy would be 100%.

Evaluations metrics are important to address the potency of the model[18], in this case, VADER. Four evaluation metrics are primarily used in classification models to get an indicator of how well the model is performing based on how it classifies, which is listed in Table 6.

Table 6. Evaluation Metrics for VADER

|  | Formula | Explanation |
| --- | --- | --- |
| Accuracy | (TP+TN)/(TP+FN+TN+FP) | The ratio of correctly predicted and total observations |
| Precision | (TP)/TP+FP) | The ratio of correctly predicted positive and total positive observations |
| Recall | (TP)/(TP+FN) | The ratio of correctly predicted observations and to all observations in the domain |
| F1-Score | (2xPrecisionxRecall)/(Precision+Recall) | Weighted harmonic mean of precision and recall. |

Note: TP = True Positive, FP = False Positive, TN = True Negative, FN = False Negative

The evaluation was conducted by comparing the results of VADER with human classified content. The sample size or support was ten tweets, and out of those, nine were labeled positive, and one was labeled neutral by the human classifiers. The VADER accuracy was 0.60, or it identified 60% of the polarity correctly as indicated in Table 7. The precision, recall, and f-score for negative and neutral were nil because the sample was not present or a correct prediction wasn't made. The model had an 86% detection rate for positive observations.

Table 7. Evaluation Metrics of VADER for Twitter

|  | precision | recall | f1-score | support |
| --- | --- | --- | --- | --- |
| neg | 0.0 | 0.0 | 0.0 | 0 |
| neu | 0.0 | 0.0 | 0.0 | 1 |
| pos | 0.86 | 0.67 | 0.75 | 9 |
| accuracy | - | - | 0.60 | 10 |
| macro avg. | 0.29 | 0.22 | 0.25 | 10 |
| weighted avg. | 0.77 | 0.60 | 0.68 | 10 |

Note: 'pos' = positive, 'neu' = neutral, and 'neg' = negative

The human-classified sample or support yielded all positive polarities and VADER was able to identify 70% of the total observations correctly (Table 8). It correctly identified all positive observations, hence the precision score of 1.00 or 100% but could not label negative classification.

Table 8. Evaluation Metrics of VADER for Reddit

|  | precision | recall | f1-score | support |
| --- | --- | --- | --- | --- |
| neg | 0.0 | 0.0 | 0.0 | 0 |
| pos | 1.00 | 0.70 | 0.82 | 10 |
| accuracy | - | - | 0.70 | 10 |
| macro avg. | - | 0.35 | 0.41 | 10 |
| weighted avg. | 1.0 | 0.70 | 0.82 | 10 |

*Note: 'pos' = positive, 'neu' = neutral, and 'neg' = negative*

The performance of VADER was less than satisfactory, albeit due to the small sample size for the comparison hence making it inconclusive to the final designation. However, it was competent in classifying positive observations correctly, with a precision score of 86% for Twitter and 100% for Reddit.

## 6. CONCLUSIONS

Contrary to the inceptive opinion that presumed cynicism, the cybersecurity content in the social media domain exhibited mostly positive or neutral sentiments. The standard and extreme negative opinions were lower than expected. Popular big techs and competent cybersecurity governments were often discussed on the platform. Due to encouraging NLP advancements, this research framework is modular and can be replicated or altered for other varying content and subject area. The accuracy of the VADER is not convincing partly due to the small sample size, but an opportunity for improvement by increasing the number of participants is possible. Future explanatory research that explains the polarity of content based on variables like length of the content, platform, the time it was posted, etc., could provide further clarity and insight.

## AUTHOR

**Bipun Thapa** is a doctoral candidate in Cybersecurity at Marymount University. His research interests are in the areas of application of Artificial Intelligence in the Information Technology space (threat classification, sentiment detection, proactive security).

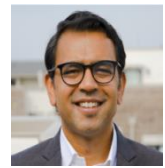